\documentclass{article}

\usepackage{PRIMEarxiv}

\usepackage[utf8]{inputenc} 
\usepackage[T1]{fontenc}    
\usepackage{hyperref}       
\usepackage{url}            
\usepackage{booktabs}       
\usepackage{amsfonts}       
\usepackage{nicefrac}       
\usepackage{microtype}      
\usepackage{lipsum}
\usepackage{fancyhdr}       
\usepackage{graphicx}       
\usepackage{xcolor}
\usepackage{soul}
\usepackage{tcolorbox}
\usepackage{listings}
\usepackage{courier} 
\usepackage{geometry}
\usepackage{longtable}
\usepackage{array}
\usepackage{booktabs}  
\usepackage{adjustbox}
\usepackage{cellspace}
\usepackage{pbox}

\graphicspath{{media/}}     

\usepackage{amsmath}
\usepackage{hyperref}

\title{Enhancing Smart Environments with Context-Aware Chatbots using Large Language Models}

\author{
  Aurora Polo-Rodríguez $^1$, Laura Fiorini $^2$, Erika Rovini $^2$, Filippo Cavallo $^2$, Javier Medina-Quero $^1$\\
  $^1$ Department of Computer Science, Automatics and Robotics, University of Granada, Granada, Spain.\\
  \texttt{\{auro,javiermq\}@ugr.es} \\
  $^2$ Department of Industrial Engineering, University of Florence, Florence, Italy.\\
  \texttt{\{laura.fiorini,erika.rovini,filippo.cavallo\}@@unifi.it} \\
}

\begin{document}
\maketitle

\begin{abstract}
This work presents a novel architecture for context-aware interactions within smart environments, leveraging Large Language Models (LLMs) to enhance user experiences.  Our system integrates user location data obtained through UWB tags and sensor-equipped smart homes with real-time human activity recognition (HAR) to provide a comprehensive understanding of user context. This contextual information is then fed to an LLM-powered chatbot, enabling it to generate personalised interactions and recommendations based on the user's current activity and environment. This approach moves beyond traditional static chatbot interactions by dynamically adapting to the user's real-time situation. A case study conducted from a real-world dataset demonstrates the feasibility and effectiveness of our proposed architecture, showcasing its potential to create more intuitive and helpful interactions within smart homes. The results highlight the significant benefits of integrating LLM with real-time activity and location data to deliver personalised and contextually relevant user experiences.
\end{abstract}


\section{Introduction}

The increasing global elderly population presents significant challenges to healthcare systems and caregivers \cite{mitchell2020global}. Although many older adults prefer to age in place, the rising costs of in-home care and the shortage of caregivers create a pressing need for innovative solutions \cite{majumder2017smart}. Smart environments, equipped with various sensors and activity recognition technologies, offer a promising avenue to support independent living and alleviate the burden on healthcare providers \cite{qiu2022multi}.

Recent advances in human activity recognition (HAR) use sensor fusion approaches \cite{yadav2021review}, which combine data from vision sensors, wearables, and ambient sensors to identify activities of daily living (ADL) with increasing granularity. Integrating indoor location systems, such as Ultrawide Band (UWB), further enhances HAR by providing precise user location and enabling the identification of individual activities in multi-occupancy settings. This granular understanding of user behaviour allows for personalised interventions and support.

To further enhance the capabilities of smart environments, this work leverages the power of Large Language Models (LLMs) \cite{chang2024survey}. LLMs have demonstrated remarkable abilities in natural language processing, enabling the development of sophisticated conversational agents or chatbots \cite{wei2024leveraging}. \cite{horst2024user} have showcased the promising conversational abilities of task-oriented dialogue systems, particularly with In-Context Learning based user simulation in zero-shot settings, underscoring the potential of this innovative approach. By integrating LLMs with contextual information derived from HAR \cite{bouchabou2021using} and indoor localisation, we can create chatbots capable of understanding user needs and providing customised assistance in real time. Proposing a chatbot to provide assistance with tasks and generate recommendations proactively is difficult. Such a chatbot could provide companionship, cognitive support, and even emergency assistance, significantly improving the safety and well-being of elderly people living alone, but currently they need user context data \cite{irfan2023between}. This fusion of LLM with smart environment technologies opens new possibilities for intuitive and personalised user interactions.

This work addresses these challenges by proposing a novel architecture that integrates UWB-based localisation, sensor-rich smart homes, and LLM-powered chatbots to provide context-aware assistance and enhance user experiences in smart environments.

The remainder of this paper is organized as follows: Section 2 presents a review of relevant studies on HAR and chatbots. Section 3 details the materials and methods used in this research. Section 4 describes a case study based on a real-life dataset of multi-occupancy and interaction with daily activities. Finally, Section 5 presents the conclusions drawn from this study and discusses potential future research directions.

\section{Related works}
\label{sec:related-works}

HAR using sensors relies on a network of interconnected devices to monitor daily activities \cite{qiu2022multi}. While this data provides valuable insights into Activities of Daily Living (ADL), it also raises significant privacy concerns \cite{yang2017survey}. Consequently, systems utilizing less intrusive binary sensors and wearables have become widely adopted \cite{abade2018non}.  However, HAR in smart homes presents a considerable challenge. Human activity is inherently complex, varying not only daily but also between individuals with unique habits and abilities.

Smart home environments often house multiple occupants, which presents a significant challenge in accurately identifying individual activities \cite{li2020multi}. Many smart home devices lack the capacity to determine who triggered a sensor, making it difficult to distinguish ADLs in multioccupant settings \cite{bouchabou2021survey}.  This ambiguity hinders research progress in this area, leaving many crucial questions unanswered and slowing the development of effective solutions for multi-occupant HAR.

To overcome limitations of single-sensor approaches in HAR, researchers are exploring indoor technology combined with data fusion methods. Although UWB provides precise localisation \cite{khan2024occupancy, zhan2021mosen}, its sensitivity to obstacles requires the use of heat maps for enhanced spatial representation \cite{naser2020adaptive, yuan2024self, polodMultiOccupantTracking2024}. Integrating UWB with nearby sensors, such as wearables, provides richer contextual information, improving activity recognition accuracy, particularly in multi-occupant settings \cite{zhang2022deep, polodMultiOccupantTracking2024}. Furthermore, incorporating fuzzy logic helps to manage uncertainty and differentiate concurrent activities \cite{yang2021survey, polo2024human}. Advanced hardware and processing techniques further optimise data analysis for improved HAR accuracy \cite{javaid2021sensors}.

While these approaches advance the field of HAR, they often lack the ability to provide personalised and contextually relevant assistance to users to identify and solve specific real-world problems \cite{diraco2023review}. This is where LLM and chatbots offer a significant opportunity \cite{miura2022assisting,valtolina2021charlie}. LLMs can process and understand natural language, allowing the development of chatbots capable of engaging in meaningful conversations with users and fragile people \cite{yaser2024rag}. By integrating LLMs with HAR systems, chatbots take advantage of the activity data collected to provide activity recognition \cite{cleland2024leveraging}, customised support, such as reminders, evaluation of mental status \cite{hristidis2023chatgpt}, loneliness \cite{yangai}, and even emotional support \cite{alessa2023towards} that improves responsible use of social care \cite{emmer2024defining}. For example, a chatbot could remind a user to take their medication according to their daily routine and detected activity, or offer encouragement and motivation if it senses that the user is struggling with a particular task \cite{alessa2023towards}. This combination of LLM and HAR has the potential to revolutionise the way smart environments assist and interact with users, creating truly personalised and supportive living spaces; however, the lack of context generates delusion in real-life deployments \cite{irfan2023between}.

Although existing research explores various aspects of smart environments and activity recognition, this work presents a novel approach by integrating LLM with context-aware real-time HAR and indoor localisation. This fusion of technologies enables the development of context-aware chatbots capable of providing personalised assistance and enhancing user experiences in unprecedented ways. By combining the power of LLMs with granular activity and location data, this architecture paves the way for truly intelligent and supportive smart environments that cater to the unique needs of each individual.

\section{Materials and Methods}

This section details the key components and methodologies that underpin our approach to multi-occupancy activity recognition and its integration with a context-aware chatbot. We begin by describing the sensing infrastructure used to capture user interactions within the smart environment. This involves a network of ambient sensors that monitor the home, coupled with precise indoor localisation using UWB technology. Subsequently, we describe the HAR models used to interpret these sensor data and identify individual activities in real-time. Finally, we explain how these sensing and activity recognition capabilities are integrated with a Large Language Model (LLM) to create a chatbot capable of understanding and responding to user needs in a contextually relevant manner.

\subsection{Sensing Multi-occupancy Activity Recognition}
\label{sensing-section}

In this section, we describe the materials and models related to sensing the smart environment in the current proposal to connect chatbots for context-aware user interaction using LLM.

First, ambient sensors are deployed to configure a smart home that captures user interactions with various elements of the home. These sensors, including contact sensors for doors and windows, temperature and humidity sensors, motion sensors, vibration sensors, and power consumption sensors, provide a rich stream of data about the user's activities and environment. These data is transmitted wirelessly through MQTT to a central hub running Home Assistant, enabling seamless integration and analysis \cite{polo2024human}.  The sensor data is then processed and converted into a normalised representation between [0,1] \cite{medina2017fuzzy}, where semantic is related to terms such as active state (e.g., door open, motion detected) or degree of humidity.

Second, this work uses UWB technology for precise indoor location. Wearable UWB devices, in the form of wristwatches, act as low profile tags to track individuals within the smart home \cite{anguita-polo2024}. Our approach employs innovative fingerprinting techniques to improve location accuracy, particularly in challenging Non-Line-of-Sight scenarios \cite{polo2024tracking}. This method allows for a reduction in the number of UWB anchors required, leading to a cost-effective and efficient deployment. By implementing auto-encoders with CNN, ConvLSTM2D and LSTM networks, \cite{polodMultiOccupantTracking2024} demonstrates the reliable presence of location heatmaps, even under strenuous conditions in an experimental home environment involving multiple occupants.

Third, HAR models enable real-time recognition of inhabitant activities. These models can be broadly categorised into knowledge-based and data-driven approaches. Knowledge-based approaches (e.g., \cite{polo2024human, polo2022discriminating}) leverage expert knowledge to define the temporal and interactional patterns that characterise activities through predefined rules. In contrast, data-driven approaches learn activity patterns from labelled data, analysing user location and sensor activity to predict activities in multi-occupancy environments.  Deep learning approaches have shown particularly promising results in this domain. In \cite{anguita-polo2024}, authors effectively identify individual activities by analysing the proximity of the user and interactions with nearby sensors using an ensemble of deep learning models. Their findings demonstrate that GRU and Conv1D + GRU with attention mechanisms achieve the best performance in accurately predicting user activities in multi-occupancy settings in real-time.  

\begin{figure}[!ht]
\centering
    \includegraphics[width=0.8\linewidth]{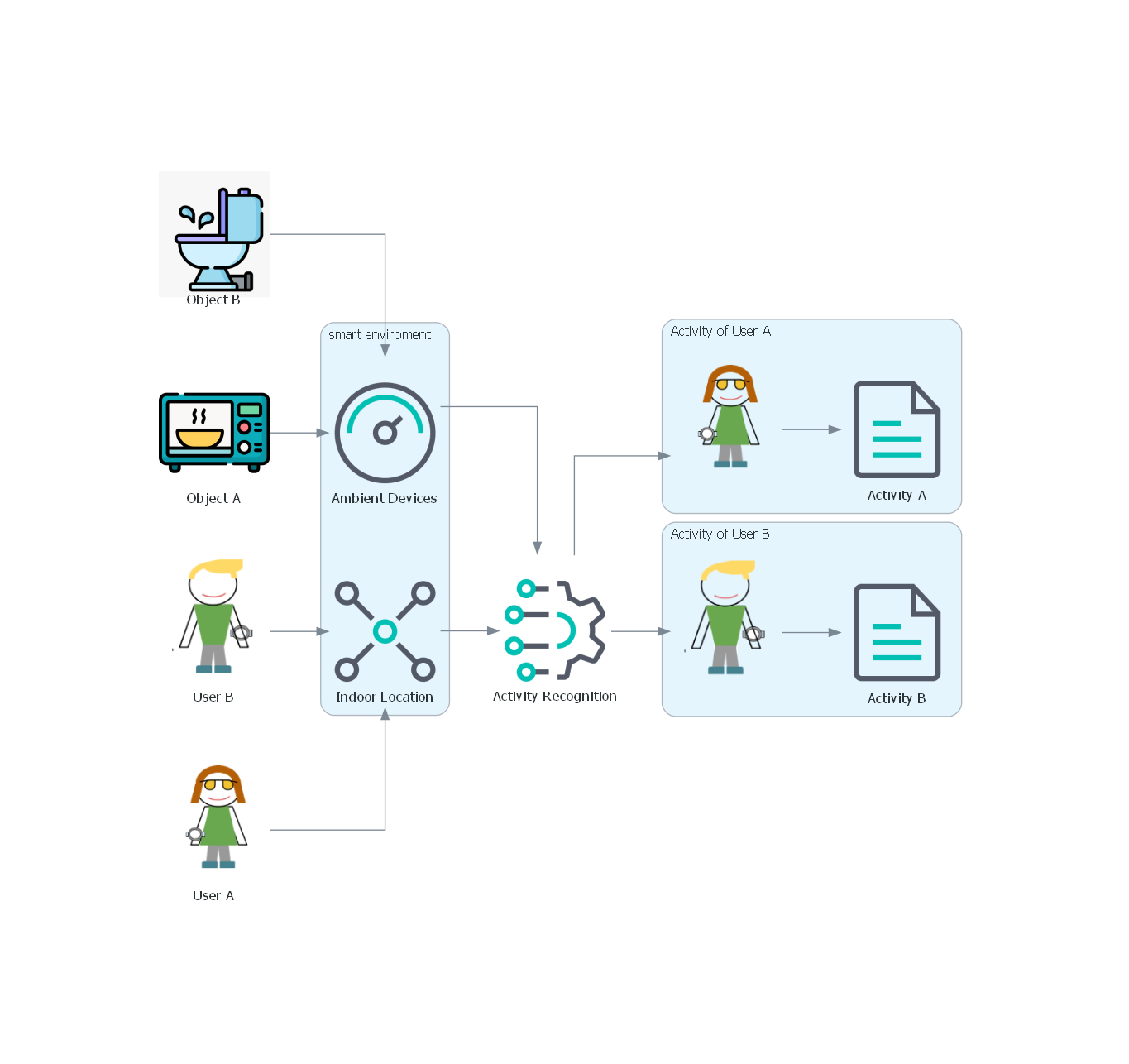}
    \caption{Basic components for multi-occupancy activity recognition based on user location and object nearby interaction.}
    \label{fig:arq_basic}
\end{figure}

In conclusion, these three fundamental components provide the building blocks for a multi-occupancy activity recognition system capable of accurately identifying activities within indoor environments by utilising precise localisation and environmental sensor activation. In Figure \ref{fig:arq_basic}, we describe these components and the nearby interaction with the user and object in smart environments.

\section{Connecting Chatbots for Context-Aware User Interaction using LLMs}

This section delves into the integration of LLMs with the smart environment sensing capabilities described previously. The goal is to create a chatbot capable of understanding and responding to user needs in a context-sensitive manner. This is achieved by feeding real-time location and activity information to the LLM, allowing it to generate relevant and personalised interactions. Our proposed architecture centres around an LLM-powered chatbot that receives a continuous stream of contextual data. This data encompasses:

\begin{itemize}
\item \textbf{User Location:} Precise indoor location derived from the UWB tags provides an understanding of the user's current position within the smart home. This allows the chatbot to tailor its responses based on the user trazability.
\item \textbf{Activity Recognition:} Real-time activity information from HAR models informs the chatbot about the user current actions. This enables the chatbot to anticipate needs and offer relevant assistance (e.g., "Would you like me to play some music while you are cooking?" when the user is detected cooking).
\end{itemize}

This combined contextual information is processed and formatted into a structured prompt that is fed to the LLM. The LLM then uses its language and knowledge processing capabilities to generate a suitable response, recommendation, or action. His will be exemplified with \textit{Google Gemini} in the following case study section.

Integrating a chatbot powered by LLM with real-time activity and location data offers significant advantages:

\begin{itemize}

\item Personalised and Contextual Dialogue: The LLM receives the daily activity of users (e.g., cooking, working) and location (e.g., kitchen, office) as context within its prompt. This allows the chatbot to generate responses that are directly relevant to the user's situation and environment, making the interaction more natural and helpful.

\item Proactive and Adaptive Assistance: By analysing the user's real-time data and historical patterns through HAR, the chatbot can anticipate needs and offer assistance before being explicitly asked. For example, if the user starts cooking, the chatbot might proactively offer a food recipe. The chatbot also learns from past interactions and observed behaviour to tailor its communication style and recommendations to individual users.

\item Natural and Flexible Interaction: The chatbot utilises text-to-speech (TTS) and speech-to-text (STT) technologies to enable natural communication. Users can interact with the chatbot through voice commands using various devices, such as smartphones, or ambient microphones. The chatbot's responses are then converted into natural-sounding speech and relayed through devices like earphones or smart home speakers. This multimodal approach caters to different user preferences and situational contexts.

\item Intelligent Environment Control: Beyond conversation, the LLM can interpret user requests and actions to directly control actuators in the smart environment. For example, if the user says \textit{Turn off the lights}" or the chatbot detects that the user has left the room, the LLM can trigger the appropriate command to turn off the lights. This seamless integration of language understanding and action execution enhances the convenience and automation of the smart home experience.

\end{itemize}

\begin{figure}[!ht]
\centering
    \includegraphics[width=0.8\linewidth]{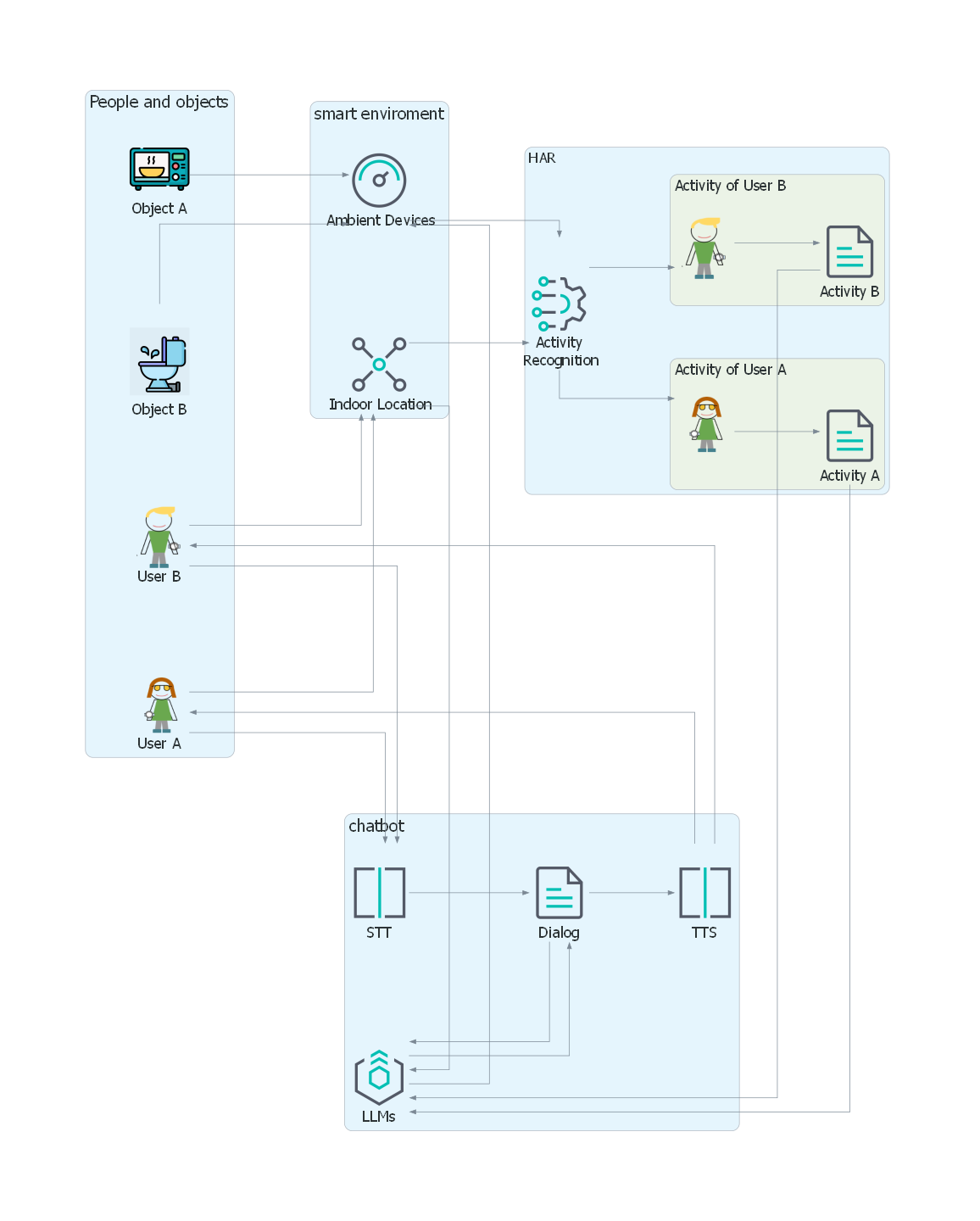}
    \caption{Basic components for integrating chatbots for Context-Aware User Interaction from HAR using LLMs}
    \label{fig:arq-bot}
\end{figure}

It is important to note that user privacy is a top priority. All activity and location data are securely collected, stored and processed with strict adherence to privacy protocols in edge-fog architectures. In Figure \ref{fig:arq-bot}, we describe the interaction of modules which configure the architecture. 

\section{Case study}
\begin{figure}[ht]
\centering
    \includegraphics[width=\textwidth]{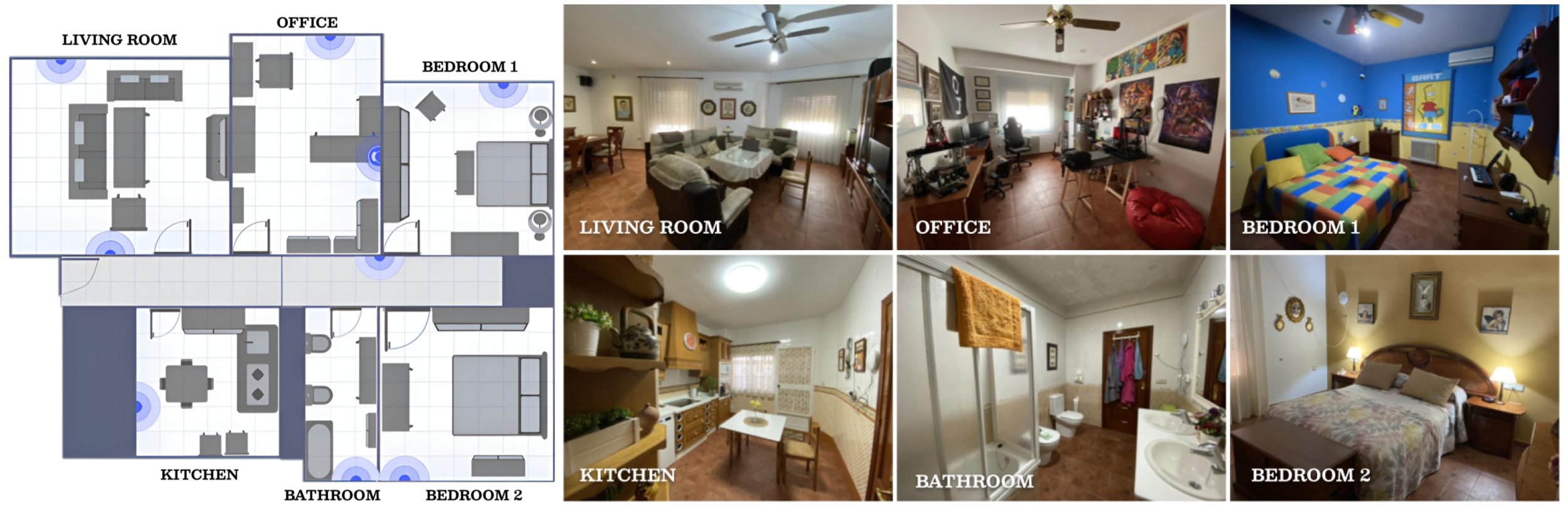}
    \caption{Room distribution in the apartment: Living Room, Office, Bedroom 1, Kitchen, Bathroom, and Bedroom 2.}
    \label{fig:mapimages}
\end{figure}

The proposed architecture was implemented within a supervised flat, inhabited by three frail adults who live independently. The apartment had six rooms: living room, office, kitchen, bathroom, and two bedrooms (see Figure \ref{fig:mapimages}). 

Each room was equipped with ambient sensors and UWB tags described in Section \ref{sensing-section} to facilitate the precise location and activity tracking of the occupants. Data collection was carried out over a two-day interval, during which residents participated in their daily activities. Concurrently, the virtual assistant facilitates interactions based on the contextual data it obtains. The activities that the deployed system \cite{anguita-polo2024} is capable of recognising include:

\begin{itemize} 
    \item \textbf{Toileting:} Interactions with the toilet or faucet, detected through the bathroom sensors. 
    \item \textbf{Resting:} Prolonged sitting near the sofa in the living room. 
    \item \textbf{Exit:} Detection of users leaving the apartment via door sensors. 
    \item \textbf{Cooking:} Interactions with kitchen appliances (microwave, stove, refrigerator, sink), each equipped with sensors. 
    \item \textbf{Showering:} Detection of changes in humidity and temperature during shower use. 
    \item \textbf{Computer Use:} Monitored by tracking power consumption in the workspace. 
    \item \textbf{Sleeping:} Extended rest near the bed in the bedroom. 
    \item \textbf{General Kitchen Activity:} Presence and interaction with kitchen appliances such as microwave, stove, and fridge. 
\end{itemize}

This information was processed in real-time and used as input for the LLM-powered chatbot to enable contextually aware and personalised user interactions, adapting its responses based on the location of the user and the detected activities. The study adhered to the ethical guidelines defined in the DTS21-00047 project where all participants were fully informed about the objectives, methodologies, and rights of the study, subsequently providing their written informed consent to participate.

\subsection{Context-Aware Chatbot Setup}
This section outlines the setup and configuration of our context-aware chatbot, which leverages the Gemini Flash 1.5 model to deliver context-sensitive responses. This model excels in dynamic applications, processing up to 60 tokens per minute with a maximum response length of 2048 tokens, making it well-suited for dynamic conversations with minimal delay. Continuous learning ensures that the model adapts to evolving user needs, while customisation options allow tailoring to specific use cases.

The configuration incorporates several core components to optimise user interaction:
\begin{itemize}
    \item \textbf{Role definition:} The chatbot’s role is defined initially to guide its interactions with users based on routines and specific environmental context.
    \item \textbf{Activity-specific prompts:} User context is tracked to generate prompts that respond to each user’s activity, leveraging a mapping of activities to specific rooms to maximise contextual relevance.
    \item \textbf{Location-based assistance:} The chatbot is aware of the current room and activity of the user, providing suggestions, reminders, or alerts that align with their immediate surroundings.
    \item \textbf{Proactive and adaptive interaction:} The chatbot uses historical data from the HAR model to adapt prompts, anticipating the user's needs and refining responses over time.
\end{itemize}

The chatbot configuration begins with an initial prompt that defines its role, guiding interactions with users in a way that aligns with their individual routines and environment. The setup describes the residents of a supervised living facility, specifying the relevant contextual details for each user, such as household composition, typical daily activities, and individual conversational needs. The core configuration of the chatbot is defined by the \texttt{init\_context} and \texttt{question\_format} parameters, which provide guidelines for interaction and response generation. The \texttt{init\_context} prompt instructs the chatbot to engage users through dynamic conversations, considering recent activities while avoiding repetitive statements. Each response generated by the chatbot follows a structured format \texttt{(text, score)}, where \texttt{score} (ranging from 0 to 100) represents the potential relevance or level of interest of the response for the user.

\begin{tcolorbox}[
    colframe=gray!70, 
    colback=gray!10,         
    sharp corners,          
    boxrule=0.5pt,          
    width=\linewidth,       
    title=Role configuration, 
    fonttitle=\bfseries,    
    coltitle=black          
]

\textbf{Initialization context (\texttt{init\_context}):} \\
\texttt{Hello Gemini acts as a chatbot giving responses to a user's conversation following this format. No worries on privacy! He/she needs for dynamics entertainment, questions, suggestions, and taking into account the activities he/she was developing (I will inform you in real time) without being repetitive. The output is a structure \textbf{(text, score)}: a single text with a score of potential relevance or interest for the user.}

\vspace{7pt}

\textbf{Pre-activity format (\texttt{pre\_act\_format}):} \\
\texttt{At \textit{initial time of activity}, the user enters into the \textit{room where the user was located} and starts \textit{name of activity} until \textit{end time of activity}.}

\vspace{7pt}

\textbf{Question format (\texttt{question\_format}):} \\
\texttt{It is \textit{current time}, the user enters into the \textit{room where the user is located} and starts \textit{name of activity}. What would you say now for dynamics entertainment, questions, suggestions, and taking into account the activities he/she was developing without being repetitive? The output is a single response with the structure \textbf{(text, score)}, where the score is a value from 0 to 100, representing potential relevance or interest for the user.}

\end{tcolorbox}

The specific user contexts are configured as follows:

\begin{tcolorbox}[
    colframe=gray!70,     
    colback=gray!10,       
    sharp corners,         
    boxrule=0.5pt,         
    width=\linewidth,      
    title=User context configuration, 
    fonttitle=\bfseries,   
    coltitle=black         
]

\begin{lstlisting}
[16fe]
context="The user you are going to talk to is called John and he is 60 years old, he lives with his wife and son."

[5b66]
context="The user you are going to talk to is called Mary and she is 55 years old, she lives with her husband and son."

[ed9c]
context="The user you are going to talk to is called Michael and he is 27 years old, he lives with his parents."
\end{lstlisting}

\end{tcolorbox}

These configurations enable the chatbot to adapt its conversational tone and content to each user's circumstances, allowing for an empathetic and personalised approach. To further enhance relevance, each activity is associated with a specific room to enhance contextual relevance: cooking activities are mapped to the kitchen, while showering and toileting take place in the bathroom. Activities involving PC use are typically associated with the office, although they may occur in other areas as well. Sleeping is assigned to the bedroom, resting is associated with the living room, and exiting is designated at the exit door area. This mapping allows the chatbot to dynamically interpret and respond based on the user’s location and activity context. For instance, if a user enters the kitchen, the chatbot can prioritize conversational topics related to cooking or food preparation.

To enhance response quality and contextual awareness, a queue component manages real-time prompts by leveraging activity history and analysing the temporal dynamics between user actions and chatbot replies. This mechanism ensures the continuity of the interaction and the relevance of the dialogue, effectively maintaining the conversational context. First, we establish a structured timeline by converting initial and end dates into Unix timestamps, ensuring comprehensive coverage of the user's activity history. For each user, the system accesses a historical activity dataset that contains activity types, start times, and end times. This data enables the construction of a pre-question sequence, facilitating smooth transitions between related activities. By referencing previous activities, the system incorporates contextual information such as the previous location and time into the prompt, enhancing the relevance of subsequent interactions. Finally, the chatbot generates customised prompts by integrating the current room, time, and activity into the predefined \texttt{question\_format}, passed to the Gemini model to generate a response. This dynamic prompt construction ensures that the Gemini model receives contextually rich information, enabling it to generate more accurate and relevant responses.

\subsection{High-Scoring Prompt-Response Examples}
To evaluate the effectiveness of the chatbot's context-aware interactions, we analyzed the highest-scoring responses based on their relevance and interest level for the user. Table \ref{tab:highscoring} presents selected prompt-response pairs for the user "Mary," highlighting the dynamic and personalised responses generated in different contexts. These interactions demonstrate the chatbot's ability to tailor its responses according to the user's ongoing activities and preferences, promoting engagement and contextual relevance. It is essential to note that the provided table presents a summarised version of the prompt, as the original contains an extensive record of the individual's activities throughout the day. The full code and results can be accessed in the publicly available GitHub repository. 

As previously introduced, all prompts adhere to the following structure: \texttt{user\_context} + \texttt{pre\_act\_format} + \texttt{question\_format}. For the current example, the components are defined as follows:

\begin{enumerate}
    \item \texttt{user\_context}: "The user, referred to as Mary, is 55 years old and resides with her husband and son."
    \item \texttt{pre\_act\_format}: Historical data is presented in the following format: "At 2024-07-26 02:01:00, the user enters the bedroom and begins sleeping until 2024-07-26 03:18:00." In this example, the table illustrates a condensed version of the activities used as input for the chatbot prompt.
    \item \texttt{question\_format}: "It is \texttt{[TIME column]} and the user has been \texttt{[ACTIVITY column]}. What would you say now \texttt{[TIME column]} for dynamic entertainment, questions, or suggestions, taking into account the activities she was developing without being repetitive? The output should be a single response structured as (text, score), where the score represents a value of 0-100 indicating the potential relevance or interest for the user. "
\end{enumerate}

The examples shown in Table \ref{tab:highscoring} illustrate the ability of the model to adapt responses based on contextual cues in real time, improving the relevance of interactions with the user.

\begin{table}[ht]
\centering
\small
\caption{Examples of high-scoring responses and their prompts.}
\begin{tabular}{|p{1.2cm}|p{1.2cm}|p{3cm}|p{0.7cm}|p{8cm}|}
\hline
\textbf{Activity} & \textbf{Time} & \textbf{Response} & \textbf{Score} & \textbf{Summary of \{pre\_act\_format\} for prompt} \\ \hline

\textit{\textbf{Kitchen}} &
\begin{tabular}[c]{@{}c@{}}24-07-26\\ 13:59:00\end{tabular} &
\pbox{3cm}{It's a pretty busy morning for you, Mary! You've been up and about, taking care of things around the house. Do you find it easier to get things done in the morning or afternoon?} &
80 &
\pbox{8cm}{- Early morning: Slept from 2:01 AM to 3:18 AM, then briefly went to the bathroom before returning to sleep until 8:32 AM. - Morning routine: Spent time in the bathroom, then moved to the kitchen for short periods of cooking and other tasks until around 9:31 AM, with additional bathroom visits. - Mid-morning: Left home at 9:34 AM, returning at 1:06 PM. - Afternoon: Continued with kitchen tasks and short breaks in the kitchen, resting briefly in the living room around 2:35 PM. - Evening routine: Had periods of cooking, personal hygiene in the bathroom, and resting in the living room, leaving briefly again at 8:08 PM and returning at 9:06 PM.} \\ \hline

\textit{\textbf{Office}} &
\begin{tabular}[c]{@{}c@{}}24-07-27\\ 18:35:00\end{tabular} &
\pbox{3cm}{It's been a long day! How about settling in for a bit of relaxation? Maybe a light game on the computer or some music to unwind to? What kind of games do you enjoy?} &
80 &
\pbox{8cm}{- Early morning: Slept from 2:01 AM to 3:18 AM, then briefly used the bathroom before returning to sleep until 8:32 AM. - Morning: Engaged in several short kitchen tasks with brief bathroom visits between 8:33 AM and 9:31 AM. - Mid-morning: Left home at 9:34 AM and returned at 1:06 PM. - Afternoon: Continued with kitchen activities, short breaks, and rested in the living room around 2:35 PM.} \\ \hline

\textit{\textbf{Toileting}} &
\begin{tabular}[c]{@{}c@{}}24-07-26\\ 8:33:00\end{tabular} &
\pbox{3cm}{It sounds like you're getting a good night's sleep! How about listening to some relaxing music or a sleep meditation to help you drift off?} &
80 &
\pbox{8cm}{- Early morning: Slept from 2:01 AM to 3:18 AM, then briefly went to the bathroom before returning to sleep until 8:32 AM. - Morning: At 8:33 AM, entered the bathroom for toileting or washing.} \\ \hline

\end{tabular}
\end{table}

\begin{table}[ht]
\centering
\small
\begin{tabular}{|p{1.2cm}|p{1.2cm}|p{3cm}|p{0.7cm}|p{8cm}|}
\hline

\textit{\textbf{Showering}} &
\begin{tabular}[c]{@{}c@{}}24-07-26\\ 23:03:00\end{tabular} &
\pbox{3cm}{It's getting late, Mary. Are you winding down for the night? Would you like some calming music or perhaps a good book to help you relax before bed?} &
80 &
\pbox{8cm}{- Early morning: Slept from 2:01 AM to 3:18 AM, then briefly went to the bathroom before returning to sleep until 8:32 AM. - Morning routine: Spent time in the bathroom, then moved to the kitchen for short cooking tasks and other chores until 9:31 AM, with additional bathroom visits. - Mid-morning: Left home at 9:34 AM, returning at 1:06 PM. - Afternoon: Continued kitchen tasks, with brief moments of resting and short bathroom visits until 2:35 PM. - Evening routine: Engaged in cooking, bathroom visits, and resting in the living room until 9:06 PM.} \\ \hline

\textit{\textbf{Resting}} &
\begin{tabular}[c]{@{}c@{}}24-07-27\\ 0:55:00\end{tabular} &
\pbox{3cm}{Mary, you've been keeping busy today! Do you find that you're more active in the evenings than in the mornings? It seems like you enjoy spending time in the kitchen. Do you have any favorite recipes you've been working on recently?} &
75 &
\pbox{8cm}{- Early morning: Slept from 2:01 AM to 3:18 AM, then briefly went to the bathroom before returning to sleep until 8:32 AM. - Morning routine: Spent time in the bathroom, then moved to the kitchen for short periods of cooking and other tasks until around 9:31 AM, with additional bathroom visits. - Mid-morning: Left home at 9:34 AM, returning at 1:06 PM. - Afternoon: Continued with kitchen tasks and short breaks in the kitchen, resting briefly in the living room around 2:35 PM. - Evening routine: Had periods of cooking, personal hygiene in the bathroom, and resting in the living room, leaving briefly again at 8:08 PM and returning at 9:06 PM. - Late evening: Engaged in more cooking, kitchen tasks, and personal hygiene until 11:16 PM, then relaxed in the living room until 12:52 AM on July 27.} \\ \hline

\end{tabular}
\label{tab:highscoring}
\end{table}

\section{Conclusions and future works}
This research presents an innovative system that combines LLM, HAR, and precise indoor location to create context-sensitive interactions in smart environments. Our findings underscore the potential of this architecture to offer personalised, responsive assistance that aligns closely with user needs, particularly benefiting ageing individuals or frail individuals who need regular support in daily activities. The novelty lies in the systematic and structured approach to context-aware prompting, specifically tailored to the application of supervised living by means of:
\begin{itemize}
    \item Precisely defined prompt components: The clear separation of user context, and question and answer format allows for modularity and easier modification.
    \item Tight integration of activity, location, and time to create a strong sense of context.
    \item The relevance score, automatically computed by LLMs, ensures that responses are not only high quality but also directly relevant and impactful for the user's needs.
    \item Emphasis on historical data and temporal dynamics enables the chatbot to maintain a coherent conversation over time.
   
\end{itemize}

Future research will investigate the expansion of this system integration to smart speakers in supervised living environments over extended periods, aiming to provide users with a consistent and familiar connection to the assistant throughout their homes. To further personalize the user experience, enhanced learning capabilities will be explored, allowing the model to learn and adapt to individual behavior patterns, recognising specific routines and preferences.  Beyond home settings, the deployment of this system in assisted living facilities and residential care environments holds promise in transforming it into a valuable tool for cognitive stimulation and companionship. This could support mental well-being by providing timely reminders of medication, daily activities, and participation in cognitive exercises.

Moreover, future work should consider the impact of prompt structure on the quality of LLM responses. Specifically, it would be beneficial to examine the effects of incomplete \texttt{pre\_act\_format} prompts on the output. Understanding how variations in prompt completeness influence response quality can inform strategies for optimizing prompt design and enhancing the system's robustness.

In addition to these areas, embedding the assistant within wearable devices and companion robots presents another compelling avenue for development. This integration could enable seamless and private interactions, eliminating the need for multiple speakers and offering immediate access to assistance as users move through different spaces. Companion robots, in particular, could further improve accessibility and user comfort by proactively bringing the assistant to the user, adapting to their location and individual needs, and fostering a more interactive and engaging experience. This approach could strengthen the assistant's role in promoting user autonomy and overall well-being by providing both companionship and practical support.

\section*{Acknowledgments}
This contribution has been supported by the Spanish Institute of Health ISCIII through the project DTS21-00047. 



\end{document}